\documentclass[lettersize,journal]{IEEEtran}

\IEEEoverridecommandlockouts   

\usepackage{cite}
\usepackage{amsmath,amssymb,amsfonts}
\usepackage{algorithmic}
\usepackage{graphicx}
\usepackage{textcomp}
\usepackage{xcolor}
\usepackage{tikz}
\usepackage{pgfplots}
\usepackage{subcaption}
\usepackage{booktabs}
\usepackage{adjustbox}
\usepackage{rotating}
\usepackage{float}
\usepackage{geometry}
\usepackage[caption=false,font=normalsize,labelfont=sf,textfont=sf]{subfig}
\usepackage{pgfplots}
\pgfplotsset{compat=1.18}

\begin{document}

\title{Torsion Resistant Strain Limiting Layers Enable High Grip Strength of Electrically-Driven Handed Shearing Auxetic Grippers} 

\author{Ian Good, Srivatsan Balaji, and Jeffrey I. Lipton
\thanks{This work was supported by the National Science Foundation, grant numbers 2017927 and 2035717, by the ONR through Grant DB2240 and by the Murdock Charitable Trust through grant 201913596.  \textit{(Ian Good and Srivatsan Balaji are co-first authors.) (Corresponding Author: Jeffrey I. Lipton.)}}
\thanks{Ian Good and Srivatsan Balaji are with the Mechanical Engineering Department, University of Washington, Seattle, WA 98195 USA (e-mail: iangood@uw.edu; sri.vatsan0307@gmail.com).}%
\thanks{Jeffrey Ian Lipton is with
the College of Engineering, Mechanical and Industrial Engineering, Northeastern University, Boston, MA 02115 USA (e-mail: j.lipton@northeastern.edu).}}%

\maketitle
\thispagestyle{empty}
\pagestyle{empty}

\begin{abstract}
Soft grippers have demonstrated a strong ability to successfully pick and manipulate many objects.
A key limitation to their wider adoption is their inability to grasp larger payloads due to objects slipping out of grasps. 
We have overcome this limitation by introducing a torsionally rigid strain limiting layer (TR-SLL).
This reduces out-of-plane bending while maintaining the gripper's softness and in-plane flexibility. 
We characterize the design space of the strain limiting layer and Handed Shearing Auxetic (HSA) actuators for a soft gripper using simulation and experiment. The inclusion of the TR-SLL with HSAs enables HSA grippers to be made with a single digit. We found that the use of our TR-SLL HSA gripper enabled pinch grasping of payloads over 1 kg. We demonstrate a lifting capacity of 5 kg when loading using the TR-SLL. We also demonstrate a peak pinch grasp force of 5.8 N, and a peak planar caging force of 14.5 N. Finally, we test the TR-SLL gripper on a suite of 43 YCB objects. We show success on 37 objects demonstrating significant capabilities.

\end{abstract}


\section{Introduction}

Soft robotic fingers have focused on emulating the ability of human and other biotas compliance when bending\cite{rus2015design}. However, the key to human’s remarkable grip is that our fingers can simultaneously bend while resisting torsion and lateral loading. People rely on a rigid skeleton with discrete joints to provide this selective compliance. We build upon a previous conference paper that introduced the torsion resistant strain limiting layer (TR-SLL) \cite{Good2024TRSLL}. The TR-SLL provides soft grippers with same torsion resistance of a skeleton without discretization. This work extends this to entirely electrically driven grippers. This allows a single Handed Shearing Auxetic (HSA) to be used in gripper and produce a high holding force. The TR-SLLs constrict bending and serves as a reaction body for the HSA. We introduce additional analysis and experiments on this new concept.

The bending of a finger defines a plane. In-plane bending and compliance is desired in a gripper. Its what allows a finger to conform to an object while still applying a normal force. Movement normal to this plane, or out-of-plane deformation, leads to grasp failures \cite{Charbel22Conformal,scharff2019reducing} and localization uncertainty \cite{Butler2019ExternalLoads}. This is especially true for antipodal or planar caging grasping where two fingers interact with an object. 

Soft robotic gripper's inability to resist out-of-plane loads has been identified as a major limiter in soft robotic systems\cite{Jain2023Reconfigurable,su2022optimizing}. 
With caging or parallel-to-gravity lifts, soft robotic systems have been able to lift car tires and dumbbells\cite{li2017fluid,li2019vacuum,li2022scaling}. However, when a gripper is perpendicular to  gravity, or when large moments are present, soft grippers lose their grasp from deformations and twists.  
This limits their ability to manipulate meaningful payloads. 

\begin{figure}[t]
    \centering
    \includegraphics[width=1\linewidth]{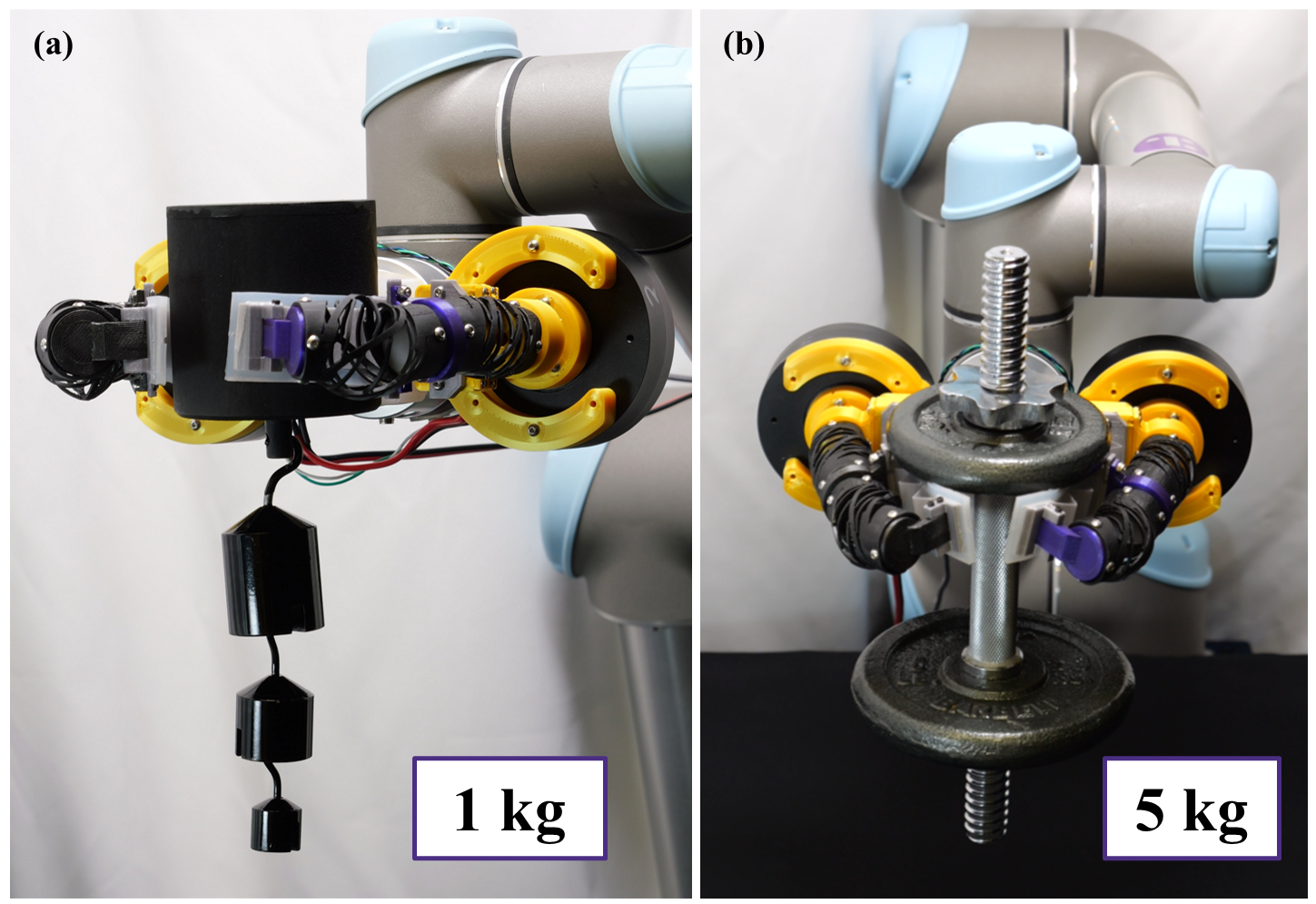}
    \caption{We create a new Torsion Resistant Strain Limiting Layer (TR-SLL) that can be added to an existing soft gripper. It increases the gripper's resistance to torsion, allowing the gripper to lift larger payloads. The skeleton of the TR-SLL can be loaded, dramatically increasing lifting capacity. Here we create a Handed Shearing Auxetic (HSA)-based gripper using the TR-SLL and two quasi direct drive motors. In (a), We show the HSA-based TR-SLL gripper lifting a 1 kg mass using a planar caging grasp. In (b), we lift a 5 kg dumbbell using the skeleton from the TR-SLL, the maximum payload capacity of the UR5 robot arm.}
    \label{fig:hero}
    \vspace{-0.2cm}
\end{figure}

This work shows how a torsion resistant strain limiting layer (TR-SLL) integrated into an electrically-driven handed shearing auxetic gripper can exceed payloads of 1 kg when grasping using the fingers as demonstrated in Fig. \ref{fig:hero}(a), and increase the lifting capacity to 5 kg when loading through the continuous skeleton as shown in Fig. \ref{fig:hero}(b).

In this paper we:
\begin{itemize}
    \item Model key parameters to understand a torsion resistant strain limiting layer
    \item Analyze the design space of handed shearing auxetics through mechanical characterization
    \item Integrate the TR-SLL with HSAs to build a soft gripper and demonstrate its enhanced payload capacity
\end{itemize}

\section{Background}

The most common place in soft robotics where strain limiting layers appear is in Pneu-nets\cite{mosadegh2014pneumatic}. The standard Pneunet uses a simple rectangular beam as the strain limiting layer (SLL). Rectangular cross sections have low second moments of inertia in one direction, allowing them to bend in the thin direction. While in theory, this would make them able to resist deformations along their thick direction, in practice, these structures have a low resistance to torsion. This causes any out-of-plane loading to generate a twist in the SLL and gripper.

For electrically-driven handed shearing auxetic grippers, a torque must be resisted to generate extension or bending of the HSA. A strain limiting layer has typically been ommited in favor of a second, opposite handed HSA \cite{chin2018compliant,chin2019simple}. In another work, rigid cable elements in addition to a second HSA are used to stiffen the system to undesired bending modes\cite{chin2020multiplexed}. The second HSA increases the moment of inertia, reducing non-planar bending modes. However, this does not reduce out-of-plane deformations as much as the TR-SLL work presented here. These works also demonstrate lower payload capacities when looking at solely HSA-based gripping. Additionally, these systems need a non-rotating grasping surface to be added to the gripper whereas the TR-SLL in this work serves both purposes.

Several strategies have emerged for improving the out-of-plane performance of soft grippers through the SLL. One approach focuses on changing the material distribution by using multi-material topology optimization\cite{wang2022topology}. Another approach discritizes the SLL and inserts hinges to allow a thicker material to still bend \cite{lotfiani2020torsional}. The first still experiences large out-of-plane deformations while demonstrating smaller increases in torsional stiffness. The second uses hinge-based strain limiting layers (SLL) which can reduce the structures ability to conform to objects compared to the continous design presented here. One method is to redesign the gripper as a whole. This can involve  putting a skeleton around the entire gripper\cite{scharff2019reducing}. However, this requires significantly more input energy to reach the same normal force output and demonstrates smaller lifting capacity. Another method involves optimizing the design of individual cells to improve torsion resistance\cite{su2022optimizing}.

Active strain limiting layers are another area of research. Many of these change their SLL stiffness through phase change \cite{zhang2019fast}, jamming chain, \cite{jiang2019chain} or through the thermoelectric properties of Field's metal \cite{buckner2019enhanced,gunawardane2022thermoelastic} or by heating hydrogels, causing them to swell\cite{visentin2021selective}. Additionally, work has been done to use jamming to increase a gripper's ability to grasp objects and increase out-of-plane grip strength\cite{crowley20223d}. 

We focus on using advancements from the field of compliant mechanisms to change the geometry of the SLL. We use triangulated beams as the torsion resistant strain limiting layer (TR-SLL) of Pneu-net. These triangulated beams resist torsion but allow bending on planar and spherical surfaces \cite{rommers2021new,naves2017building}. While one might expect the triangles along the beam to localize the deformation to interfaces between triangles, the structures deform over their entire length, with all sides of the triangle deforming when bent. Rhombus and honeycomb cross-sections have been integrated into robots using pneumatic air bladders \cite{Jiang2016Honeycomb,Jiang2021Hierarchical} and tendons \cite{childs2021rhombus} however both systems demonstrate smaller payloads than this work. The triangularized beams act as a continuum structure with enhanced stiffness to out-of-plane bending and torsion. This behaviour has a close analog with the human hand in the tendons and bones, however this work produces a continuous skeleton instead of discretized elements.

We integrate the TR-SLLs into electrically-driven handed shearing auxetic grippers and provide a step-by-step guide for integrating the TR-SLLs. Next, we perform a parametric sweep on the TR-SLLs using FEA to provide a design guide. We found that triangles close to equilateral best resisted torsion. We found in plane stress concentrations decreased with triangle count while stress from torsion followed that trend above 10 triangles per 100 mm span. We tested the integrated gripper on a cylindrical pull test and found that the addition of the TR-SLL enabled grip forces above 5 N for pinch grasps and above 14 N for planar caging grasps. Finally, we tested the gripper by picking up a weighted structure entirely using the sides of the gripper. As seen in Fig. \ref{fig:hero}, the TR-SLL gripper can pick up a 5 kg weight (the max capacity of the UR5 robot arm) using lateral loading.

Our solution allows a simple passive SLL made from a single material to be easily integrated into existing soft gripper designs. It demonstrates a large torsional spring constant and high in-plane compliance. It can be made with a low cost 3D printer and can be integrated into a variety of different soft gripper designs.

\section{Derivation of the effect of Torsion on Antipodal Gripping for Soft Bodies}

In this section we characterize the three common failure modes for soft grippers. We introduce the idea that deformations from moments are often a dominant term, and then map equations to describe this failure mode.

\begin{figure}
    \centering
    \includegraphics[width=1\linewidth]{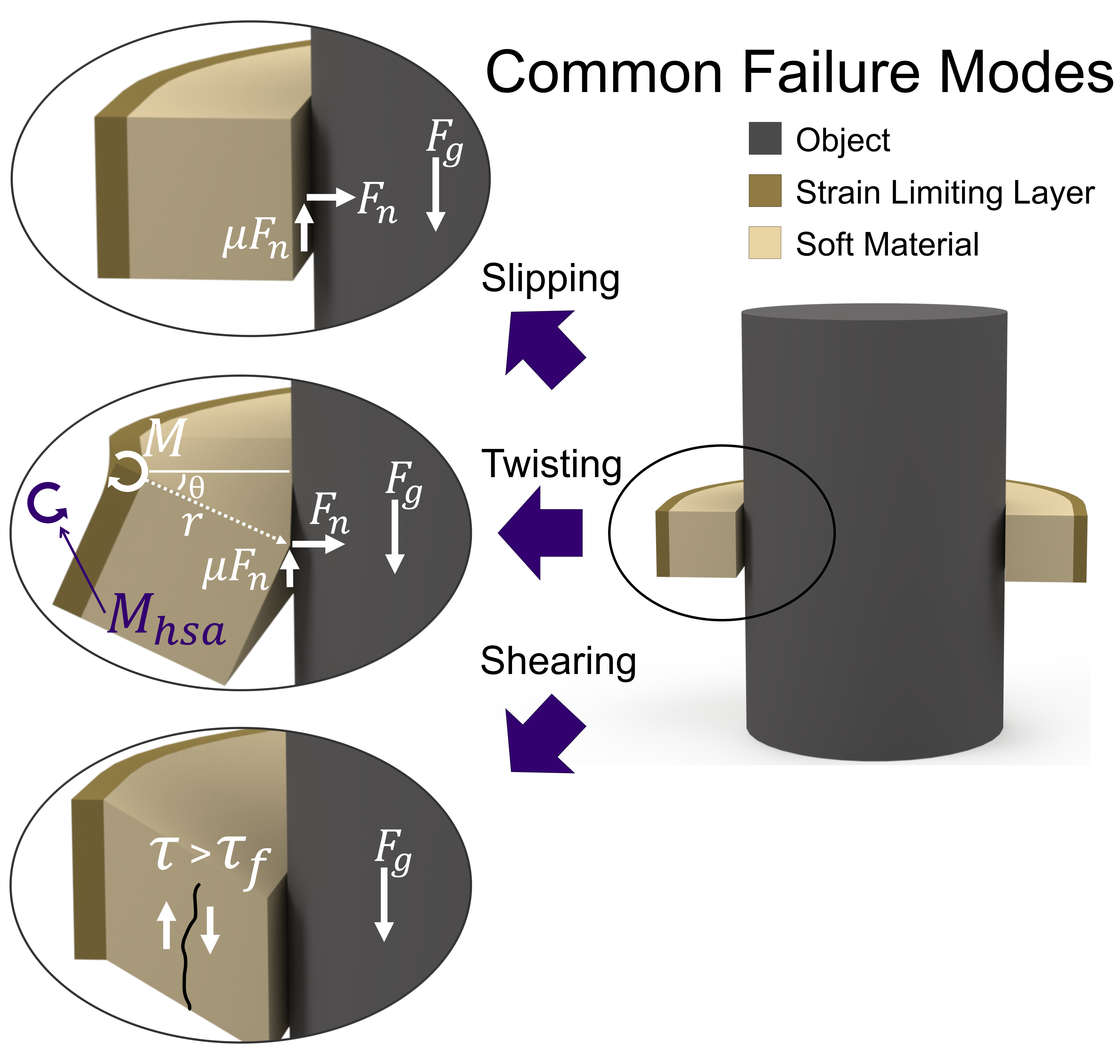}
    \caption{We demonstrate the common failure modes for soft robotic grasping. On the right we can see a simplified gripper grabbing a cylinder using an antipodal grasp. Only the Strain Limiting Layer and soft gripping material between the SLL and object is shown (not to scale). On the left, three common failure modes are shown; slipping due to low normal forces, twisting due to torsional deformation in the SLL, and shearing of the soft material. The maximum payload capacity for a this gripper is governed by the minimum of the set of these conditions. By understanding these limits, we can better inform gripper design. By including the TR-SLL, we dramatically increase our soft grippers resistance to twist. This allows it to lift larger payloads while having a better understanding of where the object is in our grasp.}
    \label{fig:failureModes}
\end{figure}

Antipodal gripping is a widespread application for soft robotic grippers. It involves picking up an object through two opposite points. There are three common limits to a soft grippers ability to perform an antipodal grasp; insufficient normal force, slippage due to torsional deformation, and shearing of the soft layer. All three of these modes as well as an example antipodal grasp can be seen in Fig. \ref{fig:failureModes}. The minimum of these three modes govern the payload capacity for an antipodally lifted object. If the payload minimum is governed by insufficent normal force, the force a gripper can lift is shown in Eq. \ref{eq:staticForce}.

\begin{equation}
     F_{static} = \mu F_n
     \label{eq:staticForce}
\end{equation}

where $F_{static}$ is the maximum force a gripper can lift, $\mu$ is the coefficient of static friction, and $F_n$ is the normal force the gripper applies as shown in Fig. \ref{fig:failureModes}. If instead the gripper is just at the limit of slip from twisting, we get:

\begin{equation}
     M = {F_gr}
     \label{eq:staticTorque}
\end{equation}

\begin{equation}
     M = \kappa\theta
     \label{eq:Torque}
\end{equation}

where $M$ is the moment on the SLL, $F_g$ is the gravitational force from the object, $r$ is the distance from the neutral axis of the gripper to the object, $\kappa$ is the torsional spring constant from twisting failure, and $\theta$ is the angle of twist relative to the normal force. These can be seen labeled in the twisting failure modes in Fig. \ref{fig:failureModes}. If we assume there is some small vertical deformation $x$ as the gripper is loaded but no slip or rolling occur, we get Eq. \ref{eq:xsmallAng}.

\begin{equation}
     x = r\theta
     \label{eq:xsmallAng}
\end{equation}

Combining Eq. \ref{eq:staticTorque}, Eq. \ref{eq:Torque}, and Eq. \ref{eq:xsmallAng} we can solve for $F_g$ as a function of $x$:

\begin{equation}
    F_g = \dfrac{\kappa}{r_{t}^2} x 
    \label{eq:forceG}
\end{equation}

Where $\dfrac{\kappa}{r_{t}^2}$ is the effective stiffness of the TR-SLL. We can use the net torque requirement from the HSA to further enhance $F_g$. We intentionally choose a HSA chirality that resists objects slipping due to the gravity field, adding to the net force supported.

\begin{equation}
    F_g = \dfrac{\kappa}{r_{t}^2} x + \dfrac{c_{\tau}}{r_{h}}  \varphi
    \label{eq:forceG+HSA}
\end{equation}
where $c_{\tau}$ is the torsional spring constant as presented in \cite{good2021expanding}, $r_{h}$ is the distance from the center of the HSA to the grasping surface, and $\varphi$ which is the angle of twist of the HSA from the motor.  The second half of Eq. \ref{eq:forceG+HSA} provides the force contribution of the HSA $F_{h}$. As shown in Fig. \ref{fig:failureModes}, the HSA's contribution is represented with the opposite direction torque $M_{hsa}$.

\begin{equation}
    M_{hsa} = c_{\tau} \varphi
    \label{eq:torqueHSA}
\end{equation}

Shearing of the soft gripping material can be another concern and is defined when:
\begin{equation}
     \tau > \tau_{f} 
     \label{eq:shearStress}
\end{equation}

where $\tau$ is the length of the shear vector in the material and $\tau_{f}$ is the shear fracture of the material. For this study, delamination was seen for high mass objects.

Together, the minimum force value from Eq. \ref{eq:staticForce}, Eq. \ref{eq:forceG+HSA}, and Eq. \ref{eq:shearStress} govern the payload capacity of a soft gripper performing an anti-podal pick on a simple object. Since soft grippers often demonstrate out-of-plane deformation as a common failure mode, this work looks to Eq. \ref{eq:forceG+HSA} as a way to increase the payload capacity of soft grippers. This paper achieves this through the creation of a continuously-deformable torsion resistant strain limiting layer. 
\section{Design of Torsion Resistant Strain Limiting Layer}

This section describes the design of our torsion-resistant strain limiting layer. First, it explores the design space for a flat strain limiting layer and evaluates them based on their bending and torsional stiffness as a function of thickness. Then, we explore the design space of the TR-SLL using FEA simulations by varying the number of triangles present over the length from two to eighty. We look at torsional stiffness, bending stiffness, as well as the average stresses present in the TR-SLL. Finally, we use the simulation results to inform which TR-SLL design to instantiate in the real world.

\subsection{Flat Strain Limiting Layers}
The first design parameter to be considered to explore the design space of flat strain limiting layers is the thickness. We chose a flat rectangular beam with a fixed length of 102 mm, and a width of 25 mm as the base of the SLL to evaluate the angular and in-plane deformation for different thicknesses. This dimension was chosen as it is the standard strain limiting layer design for soft grippers. We chose PA-6 as the base material because it has favorable elastic properties, is easy to manufacture into complex shapes, is well documented in simulation, and had similar properties as PLA which was used to fabricate the TR-SLL.

To understand the in-plane and torsional performance of the flat SLL, we analyzed the design using Ansys Mechanical 2022. We varied the thickness of the flat SLL and compared the torsional and bending stiffness. The stiffness values were calculated using the angular and in-plane deformation from the simulation. Low in-plane or bending stiffness is important as it allows less energy to be expended on complying to the desired shape. High torsional stiffness is important because a major failure mode for soft gripper not being able to lift an object is due to twisting of the grasp. Large values of torsional stiffness can cause the grasp failure condition to switch from twisting to a lack of normal force or shearing of the soft material. Additionally, high torsional stiffness allows for better reasoning over where objects are in the grasp.

The data from the FEA analysis of the flat strain limiting layer can see seen in Table \ref{tab:ssl-analysis}. We can see that torsional and bending stiffness increase as the SLL thickness is increased. The performance ratio of in-plane displacement over angular displacement also improves as thickness decreases. We apply a moment that is an order of magnitude smaller than the moment applied to the TR-SLL due to its reduced torsional stiffness, and to achieve reasonable deformation for the flat SLL.

\begin{table}[t]
\caption{Simulated Torsional and Bending Stiffness for the Flat Strain Limiting Layer (SLL). Fixed dimensions of the SLL are 102 $\times$ 25 $(l \times w)$ in mm.}
\begin{tabular}{ccc}
\hline
\begin{tabular}[c]{@{}c@{}}Thickness\\ {[}mm{]}\end{tabular} & \begin{tabular}[c]{@{}c@{}}Torsional Stiffness [Nmm/rad] \\ (M = 0.5 Nmm) \end{tabular} & \begin{tabular}[c]{@{}c@{}}Bending Stiffness [N/mm]\\ (F = 0.01 N) $\times 10^{-3}$\end{tabular} \\ \hline
0.3 & 1.04 & 0.23 \\
0.4 & 2.30 & 0.47 \\
0.5 & 4.44 & 0.87 \\
0.6 & 7.62 & 1.47 \\
0.7 & 12.05 & 2.23 \\
0.8 & 17.93 & 3.46 \\
0.9 & 25.46 & 4.93 \\
1.0 & 34.81 & 6.67 \\ \hline
\end{tabular}
\label{tab:ssl-analysis}
\end{table}

\subsection{Simulation of Torsion-Resistant Layer Strain Limiting Layer (TR-SLL)}
\begin{figure*}[t]
    \centering
    \includegraphics[width=1\textwidth]{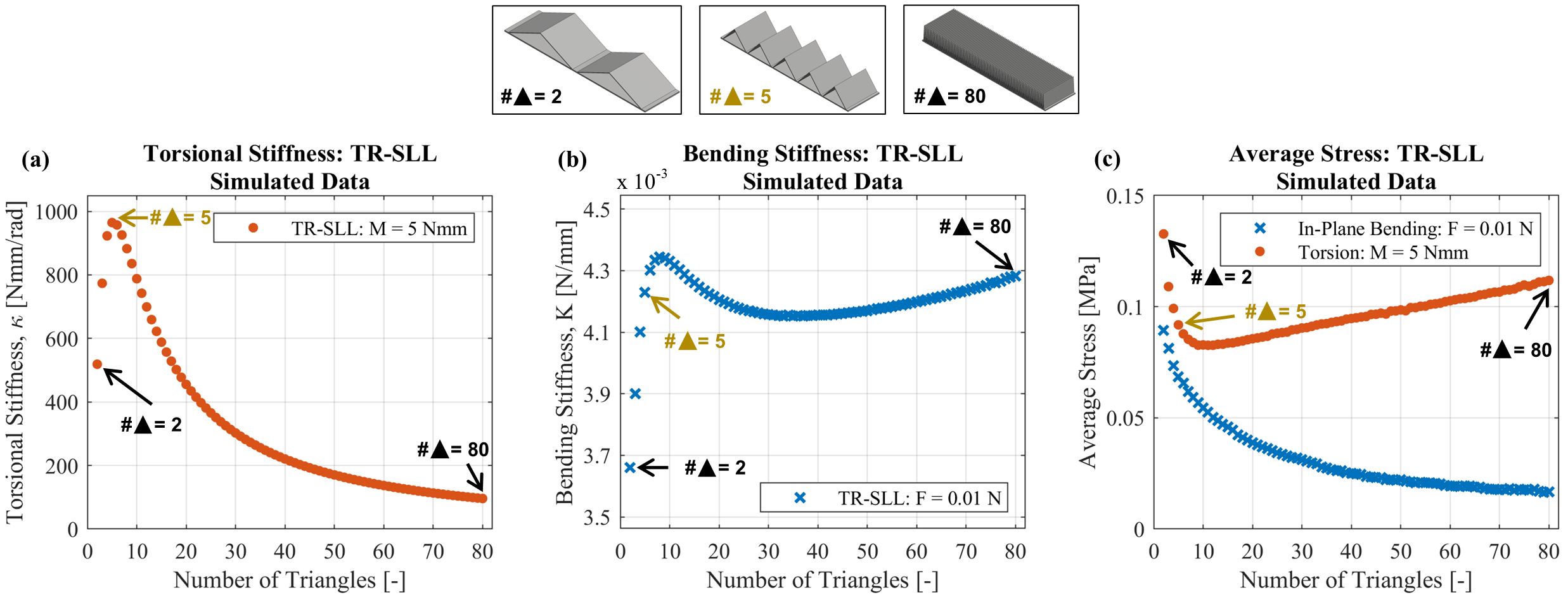}
    \caption{Plot of (a) torsional stiffness, (b) bending stiffness, and (c) average stress during bending and torsion as a function of number of triangles in the TR-SLL. In (a), we see torsional stiffness of the TR-SLL calculated from out-of-plane deformation due to an applied moment of 5 Nmm. The most torsionally resistant design is the one with five triagnels. In (b), we see the bending stiffness calculated from in-plane deformation due to an applied force of 0.01 N. The design that uses the least amount of input force to achieve the same displacement is the two triangle design. The average stress data in (c) provides insights on stress concentrations in the TR-SLL. All stress values are well below the yield of the material tested.}
    \label{fig:triangleFEAresults}
\end{figure*}

The next step is to analyze the design space of the TR-SLL. We select a number of triangles as a parameter to explore the design space. We fixed the length of the TR-SLL base to be equal to 102 mm and the height to be 12 mm. 12 mm was chosen as a reasonable value to search the design space without significantly increasing the volume of the gripper, and to easily retrofit onto pre-existing works. For the base of the TR-SLL, we included a 1-mm long extension at the ends of the TR-SLL to ensure proper meshing around the area where the triangles meet the base. This gives 100 mm for complete triangles to be generated over. Because our design relies on having complete triangles, this discritizes our search space to be angles that result in integer numbers of triangles along the length of the TR-SLL. The number of triangles was varied from two to eighty. Examples of three such configurations can be seen in Fig.  \ref{fig:triangleFEAresults}. Due to manufacturing limitations, the smallest realizable beam thickness as tested was 0.4 mm. This was used for the thickness of the triangles, however 0.8 mm (double the realizable thickness) was used as this provided a continuous path for the 3D printer nozzle to follow. Alternative manufacturing methods like resin printing or casting could eliminate this restriction. 

In-plane and torsional bending conditions for the TR-SLL were simulated using Ansys Mechanical 2022 FEA software. Similar to the flat SLL, PA-6 was used as the model material. A tetrahedral mesh with an element size of 0.2 mm was chosen to achieve consistent results for all the design configurations. For out-of-plane twisting, one end of the TR-SLL was fixed, and a bending moment of 5 Nmm was applied to the free end of the beam. This simulates the twisting failure condition that occurs during object grasping. Similarly, for in-plane bending simulation, one end of the TR-SLL was fixed and a force of 0.01 N was applied to the free edge of the beam. This simulates the condition that occurs when the HSAs are driven and cause bending. The total deformation and equivalent stress values are obtained from the simulation, and the deformation values are used for calculating stiffness for in-plane bending and out-of-plane twisting conditions.

The torsional stiffness values start to rise from two triangles, and then reaches a maximum at five triangles before dropping as the triangle count increases. The eighty-triangle design configuration had the least amount of torsional stiffness, meaning that it has the highest angular deformation of all TR-SLL designs. Beyond 30 triangles, the slope of the curve flattens considerably. A similar trend holds true for the bending stiffness plot. The two triangle design results in a local minima that rises to a maximum for the eight-triangle design before continuing to fall as triangle count increases. The curve flattens, and then starts to rise beyond the 33-triangle design. As predicted by statics, designs around the equilateral triangle best resist deformations.

Using linear elastic strain energy theory, we can compare the rotational stiffness of each of the beams following:

\begin{equation}
     \kappa = \frac{M}{\theta} 
     \label{eq:rotationalstiffness_LESE}
\end{equation}

where $\kappa$ is the torsional stiffness, $M$ is the applied moment, and $\theta$ is the angular displacement. The value for the 0.4 mm SLL in Table \ref{tab:ssl-analysis} result in a torsional stiffness of 2.3 Nmm/rad while the values from Fig. \ref{fig:triangleFEAresults}(a) result in 518.7 Nmm/rad for the two-triangle TR-SLL, and 964.76 Nmm/rad for the five triangle TR-SLL. The inclusion of triangles to the SLL has a dramatic impact on torsional resistance. Since torsional stiffness directly contributes to the twisting failure mode, we choose the TR-SLL with the highest resistance to torsion to design a gripper, as this minimizes the twisting failure. This TR-SLL design also had lower average stress while twisting, as it can been in Fig. \ref{fig:triangleFEAresults}(c). Since the applied loads are relatively small, the stress values are correspondingly small. These values are well below the failure criteria for the materials (PA-6 or PLA), but they serve as helpful guides to trends in stress values as the number of triangles is changed and as applied loads are increased.

\begin{figure}
    \centering
    \includegraphics[width=0.48\textwidth]{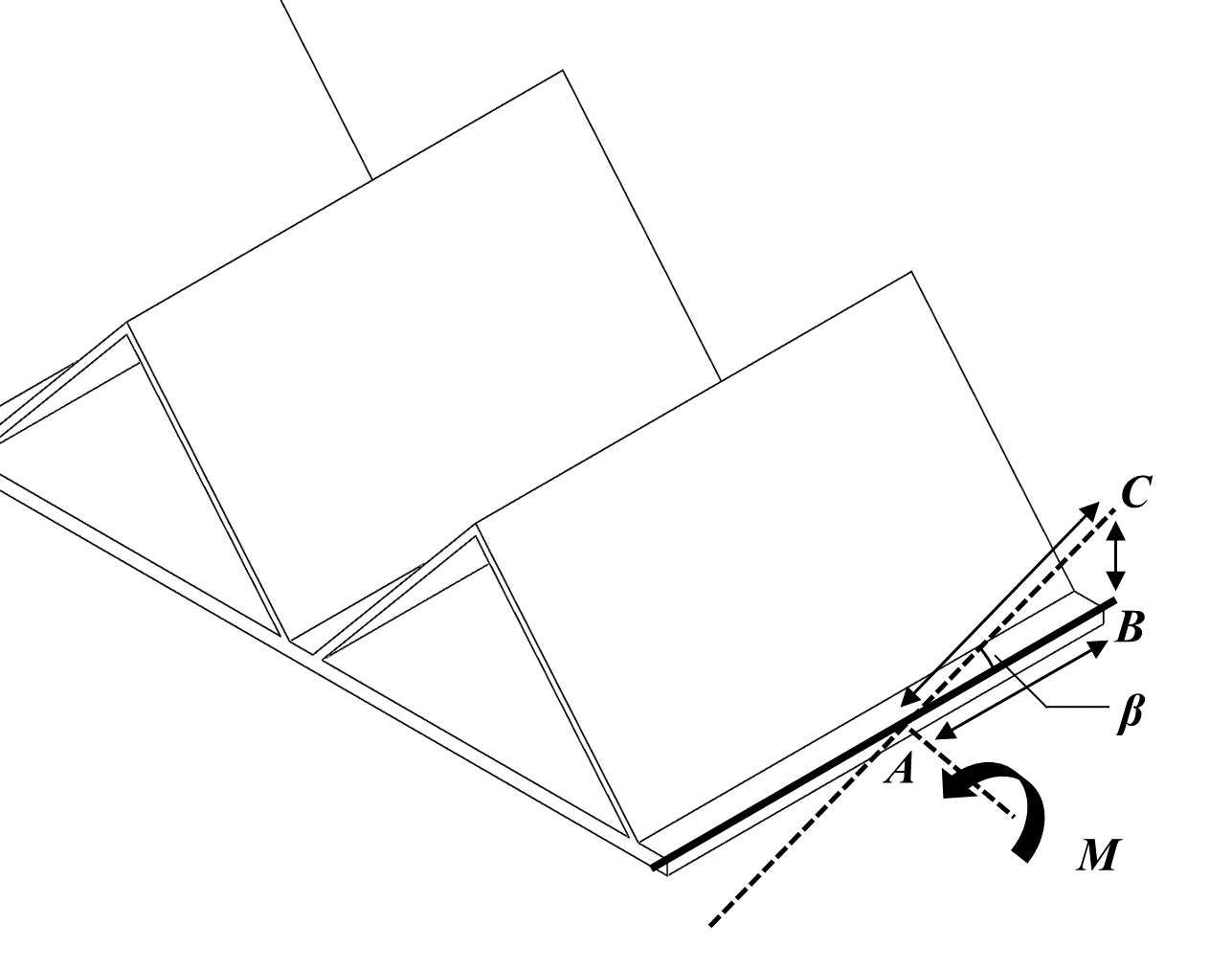}
    \caption{Torsional bending setup for simulated angle measurement. The example shown here is a five-triangle TR-SLL. A moment M is applied about the longitudinal axis of the TR-SLL. Line AB is deformed to line AC forming the angle $\beta$, which is then calculated using Eq. \ref{eq:angle-calculation}. The number of triangles was varied across a span of 100 mm resulting in different triangle widths.}
    \label{fig:angle}
\end{figure}

Torsional stiffness was calculated for each TR-SLL configuration for an applied moment of 5 Nmm using the angle of rotation and Eq. \ref{eq:rotationalstiffness_LESE}. For calculating the angle of rotation, we employed Eq. \ref{eq:angle-calculation} to find the angle $\beta$ using the deformation values from the simulation. 
\begin{equation}
    \beta = \sin^{-1}\bigg(\dfrac{BC}{\sqrt{(AB)^2 + (BC)^2}}\bigg)
    \label{eq:angle-calculation}
\end{equation}

\noindent where, AB is the distance from the center to the edge, BC is the linear displacement in vertical direction, $\beta$ is the angle between line AB and AC, as represented in Fig. \ref{fig:angle}.

Similarly, the in-plane deformation values were used to calculate the bending stiffness for an applied force of 0.01 N for each configuration. The stiffness values were plotted against the corresponding number of triangles that were present in the TR-SLL, and the results are shown in Fig.  \ref{fig:triangleFEAresults}.

\section{Fabrication of Handed Shearing Auxetic Gripper}
In this section, we describe the manufacturing and characterization of the Handed Shearing Auextic (HSA) actuators, and the fabrication and assembly steps for the gripper. We present the fabrication process for the TR-SLL, and describe the test methods to evaluate the performance of our gripper.

\subsection{Characterization of Handed Shearing Auxetics (HSA)}
\begin{figure*}[t]
    \centering
    \includegraphics[width=1\textwidth]{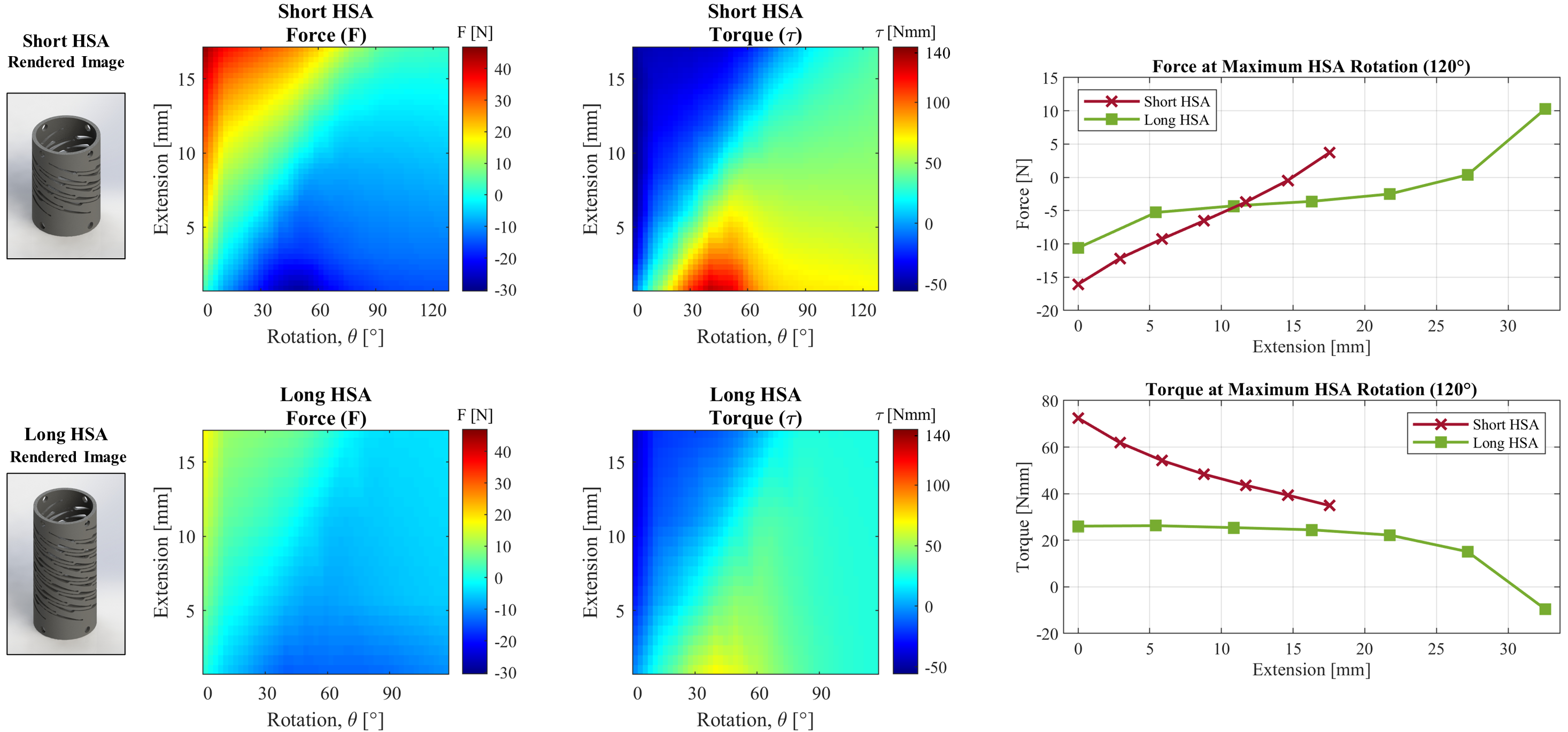}
    \caption{This presents the linear interpolated performance heat maps of the short and long HSAs. We measure the mean force and torque as a function of extension and rotation. Negative values represent the HSA pushing on the environment, and positive values represent it pulling on the environment. In the heatmap we see non-linear behavior with a maximum negative force occurring around 0 mm extension and 45$^\circ$ for both HSAs. Values start to shrink beyond this as the HSA experiences larger deformations. Torque values follow a similar trend. Slices of the heatmaps are plotted for the maximum HSA rotation. This corresponds to the angle used when lifting objects. In the sliced 2D plots, we see the force (vs) extension, and the torque (vs) extension data for the long and short HSAs at 120$^\circ$ rotation.}
    \label{fig:performanceheatmap}
\end{figure*}

In order to establish a design for our gripper, we characterized HSAs using two main properties, force (F) and  torque ($\tau$) as a function of both displacement and rotation. We characterized two different configurations of the HSA, long and short. The long HSA is described as the configuration with eight rows and five columns of unit cells. This configuration provides an extension that allows the gripper to fully close. The short HSA has four rows and five columns of unit cells. The short HSA is included in this work as the long HSA would be poorly constrained with just a connection at the proximal and distal ends. All the HSAs we tested had an outer diameter of 31.75 mm and a wall thickness of 1.8 mm. Rendered images of the two configurations of the HSAs can be seen in Fig. \ref{fig:performanceheatmap}.

When selecting HSAs for the gripper, a connection at the midpoint of the TR-SLL is required. This ensures that the HSA extension couples directly with TR-SLL bending. For the gripper, two short HSAs are used. 

All HSAs we tested were fabricated using the Carbon M1 3D printer with FPU 50. All HSAs were printed horizontally on the build platform, and were post processed following the manufacturer's guidelines and specifications.

The test procedure consisted of a series of extensions and rotations that were programmed using a custom test method on an Instron 68SC-2 at 200 Hz. We varied the extension from 0 to 30 mm, and the rotation from 0 to 120$^\circ$. The ranges of extension and rotation were determined by manually observing them for the two configurations. The data was collected over seven and fourteen equidistantly spaced points in displacement and rotation respectively. This results in 98 datapoint groups per HSA. Each datapoint group was collected during a 1 mm vertical displacement at 10 mm/s. The raw data collected from the test apparatus was approximated using cubic fits for displacement and rotation to visualize the force (F) and torque ($\tau$) responses of the HSA as shown in Fig. \ref{fig:performanceheatmap}.

A linear interpolation surface fit was chosen to visualize the data. For the force plot, positive values represent the HSA pulling on the environment while negative values represent pushing on the environment. For torque, positive values represent the HSA trying to twist less in the direction it is twisting and negative values represent the HSA trying to twist more than the direction it is twisting in. When added to the gripper, positive torque values positively contribute to payload capacity from the perspective of Eq. \ref{eq:forceG+HSA}. 

For the short HSA, force values reach a maximum of 45.9 N at 17.1 mm extension and 0.0$^\circ$ rotation and a minimum of -29.6 N at 0.0 mm extension and 47$^\circ$ rotation. For torque values, the short HSA reaches a maximum of 139.2 Nmm at 0.0 mm extension and 39.2$^\circ$ rotation and a minimum of -56.02 Nmm at 8.95 mm extension and 0.0$^\circ$ rotation.

For the long HSA, force values reach a maximum of 23.56 N at 31.9 mm extension and 0.0$^\circ$ rotation and a minimum of -13.55 N at 0.0 mm extension and 48.4$^\circ$ rotation. For torque values, the long HSA reaches a maximum of 68.6 Nmm at 0.0 mm extension and 38.7$^\circ$ rotation and a minimum of -41.2 Nmm at 15.9 mm extension and 0.0$^\circ$ rotation. 
This work uses a bang-bang position command for opening and closing the gripper. So we compared the performance of the two HSA designs by looking at force and torque at 120$^\circ$ HSA rotation as a function of displacement. This makes it easier to visually understand what is happening during grasp.

\subsection{Fabrication Process}
\label{fab}
\begin{figure}
    \centering
    \includegraphics[width=0.48\textwidth]{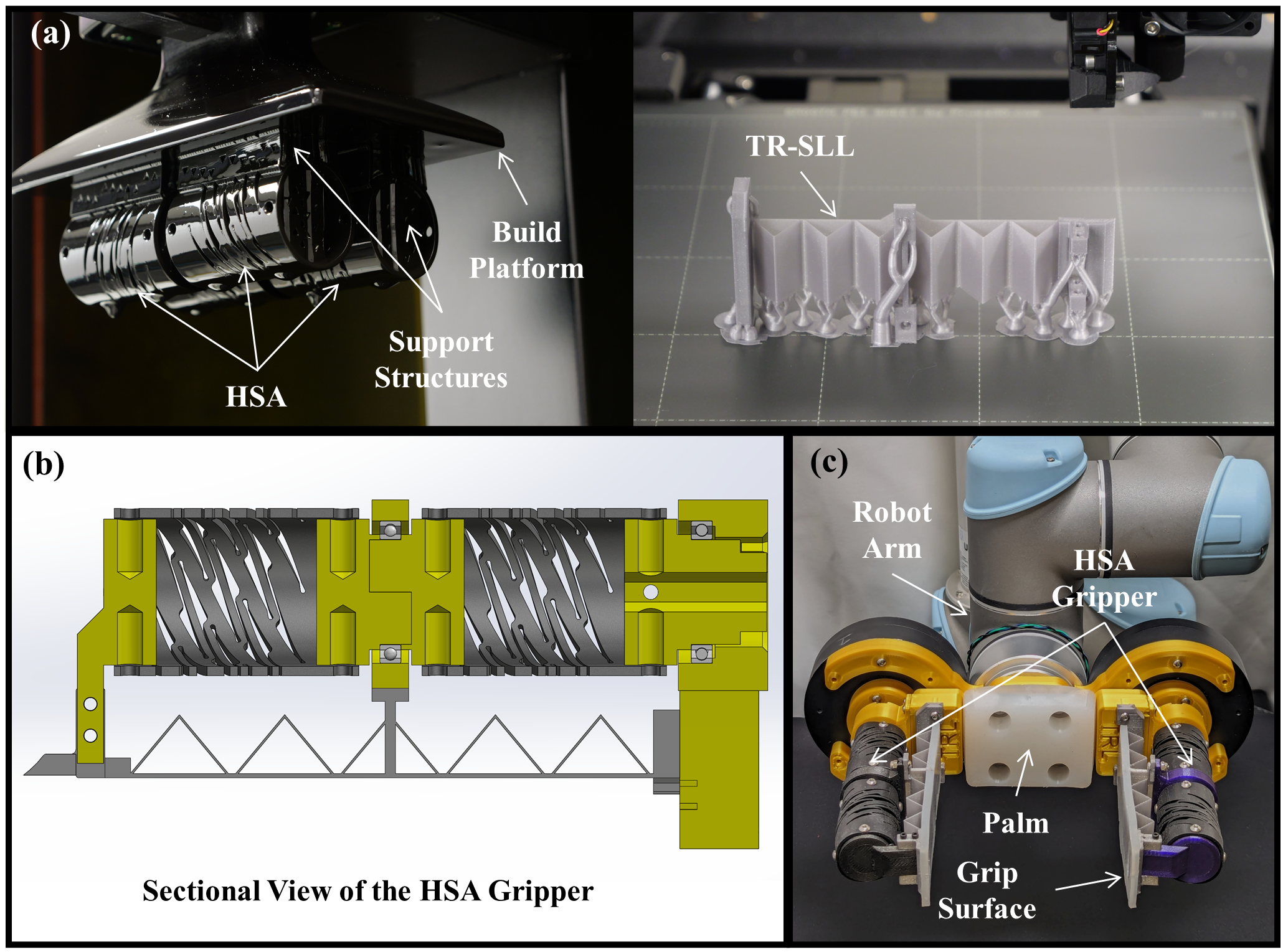}
    \caption{This figure shows the fabrication processes of the HSA gripper. First, (a) we print the HSAs using the stereolithography 3D printing process with FPU 50. We also print the TR-SLL with PLA using a standard FDM printer. For the TR-SLL, we cast a layer of Ecoflex 00-30 to achieve a high coefficient of friction. Then in (b), we assemble the components with adaptors and bearings as shown. Finally, we mount the gripper to a UR5 robot arm with two quasi direct drive motors using a shaft as shown in (c). We also mount a cast Ecoflex 00-30 palm to perform caging grasps. The total gripper weight is 1521 g with motors and 386 g without motors.}
    \label{fig:gripper-fabrication}
    \hfill
\end{figure}

The fabrication process for the HSA gripper can be seen in Fig. \ref{fig:gripper-fabrication}. First, the HSAs and the TR-SLL were fabricated using SLA and FDM printing processes respectively. Here, we use two left-handed short HSAs for one finger, and two right-handed short HSAs for the other finger. The HSAs were 3D-printed out of FPU 50, while the TR-SLL was printed out of PLA. Additionally, all of the required components, adaptors, and mounts of the HSA gripper were 3D-printed using the FDM process with PLA. The middle section of our gripper contains a bearing that allows rotation of the HSAs, while acting as a constraint for the TR-SLL. At the proximal end of the gripper, a bearing allows the HSAs to be rotated by the motor. We add a layer of Ecoflex 00-30 to the TR-SLL's base and to the gripper's palm to increase the coefficient of static friction. The final assembly of the HSA gripper can be seen in Fig. \ref{fig:gripper-fabrication}. Once assembled, the gripper is mounted on the UR5 arm using four bolts.
The gripper weighs 1521 g with the motors and 386 g without the motors and their mounts. If a low inertia end-effector is desired, a flex shaft can be used to transmit motor torque to the gripper.

\section{Results - Performance Evaluation of HSA Gripper} 

In this section, we evaluate the performance of our HSA gripper. We test the HSA gripper using the grip force characterization, where the gripper grabs a 90 mm diameter cylinder that is pulled out of its grasp. This demonstrates the maximum payload capacity for the grippers. Finally, the HSA gripper is tested on a subset of the YCB dataset. The subset features objects that fit within the HSA gripper with pinch and planar caging grasps. We show an overall success rate of 86.\% out of the 43 objects we tested.

\subsection{Grip Slip Force Characterization}

\begin{figure}
    \centering
    \includegraphics[width=0.48\textwidth]{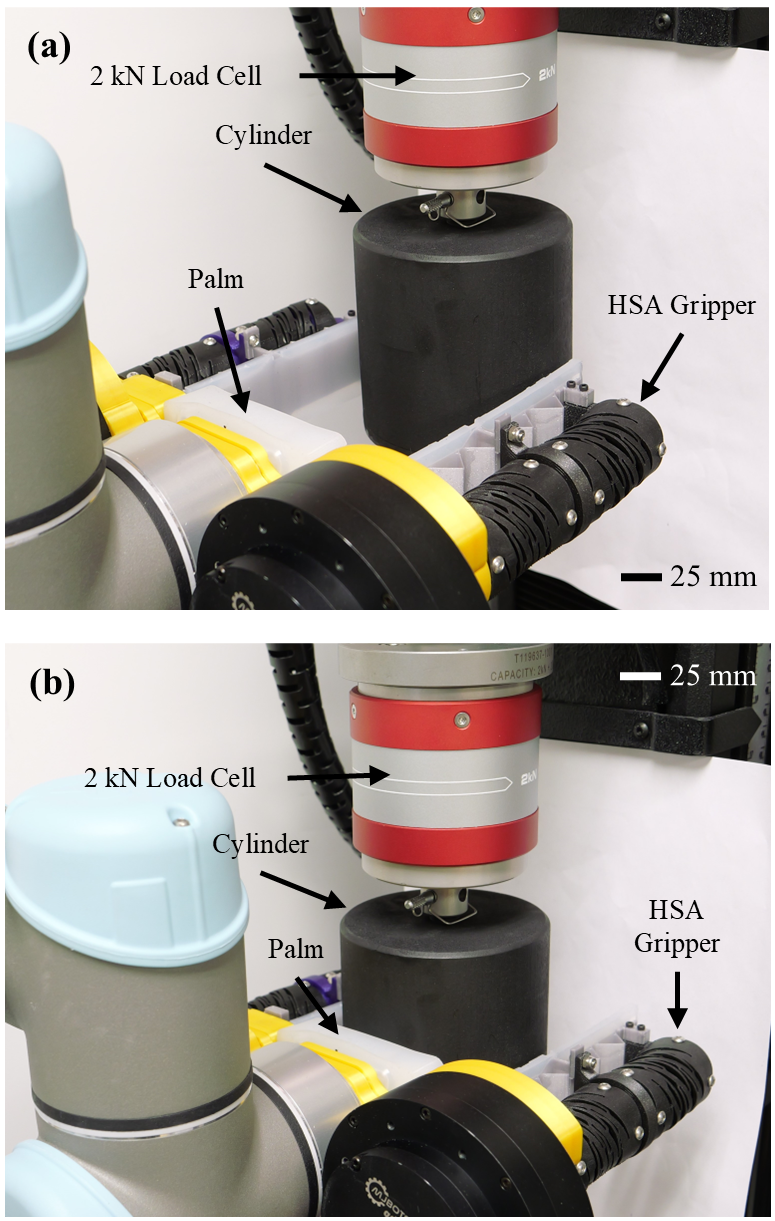}
    \caption{Here the Instron testing setup for HSA gripper characterization is presented. The HSA grippers and the motors are fixed to a UR5 robot arm. They are positioned to perform (a) antipodal and (b) planar caging grasps using a painted-, 3D-printed cylinder that is 90 mm in diameter. The HSAs were rotated 120$^\circ$ to grasp the test cylinder. Then, the cylinder is forced down and the force response is collected by the transducer. This demonstrates the minimum of the three common antipodal failure modes for soft grippers; slipping, twisting, and shearing.}
    \label{fig:load-capacity}
\end{figure}

\begin{figure}
    \centering
    \includegraphics[width=0.48\textwidth]{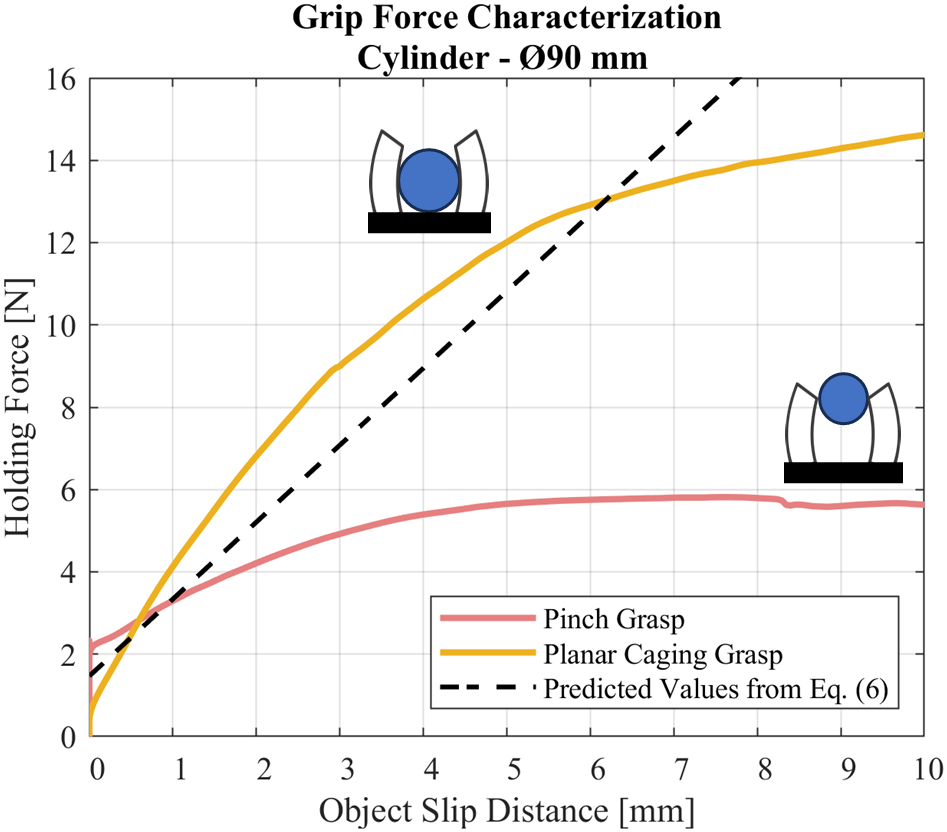}
    \caption{This figure shows the data from the grip force characterization test for two grasping modes, pinch and planar caging, and the predictions using Eq. \ref{eq:forceG+HSA}. The TR-SLL gripper achieved a peak holding force of 5.8 N for the pinch grasp, and a peak holding force of 14.6 N using planar caging grasp. The larger grip stiffness demonstrates a higher minimum antipodal grasp failure condition and can allow for better understanding of an object's position when grasped.
    }
    \label{fig:instronResults}
\end{figure}

We characterized the grip strength of our TR-SLL gripper using the grip force test. This test demonstrates the minimum of the three common antipodal failure modes for soft grippers; slipping, twisting, and shearing. Data was gathered using an Instron 68SC-2 at 200 Hz, and the test setup can be seen in Fig. \ref{fig:load-capacity}. Pinch and caging grasps were performed around a painted, 3D-printed 90 mm diameter test cylinder made from VeroWhite on a Stratasys J750 Digital Anatomy printer. The grippers were mounted to a Universal Robotics UR5 arm. The HSAs were rotated 120$^\circ$ to grasp the cylinder. The cylinder was then moved downwards at a rate of 1 mm/s for 10 mm. The force contributions of the HSAs and the TR-SLL were recorded, and the data from the experiment can be seen in Fig. \ref{fig:instronResults}.

The prediction from Eq. \ref{eq:forceG+HSA} is also shown in Fig. \ref{fig:instronResults}. To fully analyze this equation, we need use the data for $\kappa$ for the five triangle TR-SLL as shown in Fig. \ref{fig:triangleFEAresults}. We measure $r_t$ and $r_{h}$ using both SolidWorks and micrometers to record the exact thickness of the Ecoflex grip surface using a median measurement. $x$ is recorded during the Instron tests which leaves $c_{\tau}$ to be calculated.

\begin{figure}
    \centering
    \includegraphics[width=0.48\textwidth]{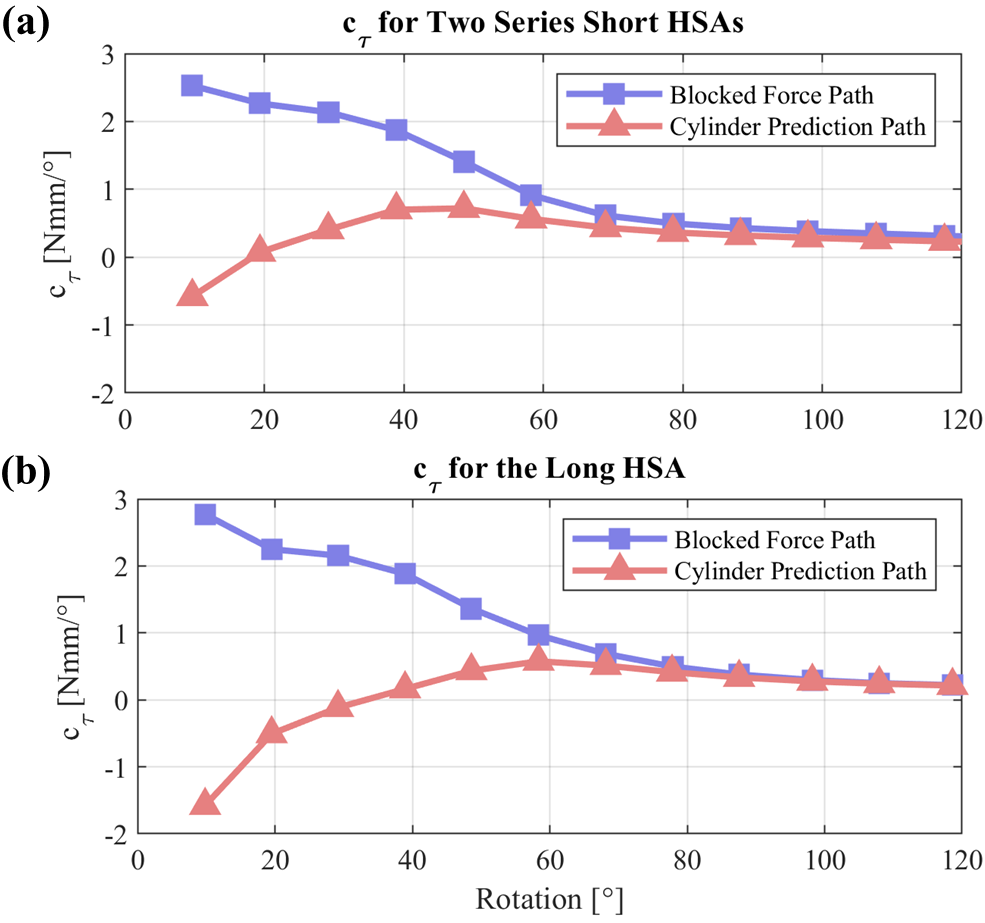}
    \caption{This figure represents predicted torsional spring constant $c_\tau$ as a function of HSA rotation for the short (a) and (b) long HSAs.}
    \label{fig:c_tauPredict}
\end{figure}

There are two values of $c_{\tau}$ that can be modeled. We can consider two short HSAs in series, or use the data from the long HSA. Additionally, the HSA displacement must be considered. For this work, we compute $c_{\tau}$ for both the blocked force path (i.e. the path that allows zero extension and only rotation) as well as the path that accounts for the gap between the gripper and the object as it pinches the cylinder. We call the former the blocked force path and the latter the cylinder prediction path. All four cases are modeled and presented in Fig. \ref{fig:c_tauPredict}(a) and \ref{fig:c_tauPredict}(b). Raw data from the Instron tests were used to compute the values of $c_{\tau}$ for the first two cases. The blocked force path represents an upper absolute bound on values for $c_{\tau}$ for the HSAs presented here as the maximum positive torque value occurred with no displacement as seen in Fig. \ref{fig:performanceheatmap}.

The long HSA is selected for analysis in Fig. \ref{fig:instronResults} since it does not require the assumption of two identical series springs. Since the grip force characterization uses the cylinder, the cylinder prediction path data is used. Since $c_{\tau}$ is calculated by dividing the torque by the rotation, values around 0$^\circ$ rotation tend towards infinity. For this work, we operated the HSAs using a bang-bang controller so all grasps were tested at 120$^\circ$.

This work also predicts the torsional spring constant $c_{\tau}$ by looking at the applied force from a finger tip.
The finger is placed a certain distance from a torque cell and then the finger is then commanded closed such that it always touches the torque cell at the same distance. This test was repeated at different gripper to torque cell gaps. The following distance were considered for our measurement: 0 mm (contact), 10 mm, 20 mm, 30 mm and 40 mm. The mean applied normal force was calculated using the raw torque data collected over a period of 10 seconds. We then established a relationship between the normal force and torsional spring constant of the HSA by balancing the force components in the system. We know that:

\begin{equation}
    F_{h} = F_g - \mu F_n
\end{equation}

Here, $F_n$ is the force component of the HSA from Eq. \ref{eq:forceG+HSA}. We know that $\mu F_n$ represents the static force in the system from Eq. \ref{eq:staticForce}. The coefficient of static friction $\mu$ was experimentally determined to be 1.27 for Ecoflex on a painted 3D-printed Vero material. By rearranging Eq. \ref{eq:torqueHSA}, we can predict torsional spring constant $c_{\tau}$ using the following equation:

\begin{equation}
    c_{\tau} = \dfrac{F_{h} * r_{h}}{\varphi}
\end{equation}

The predicted values of $c_{\tau}$ from the measured normal force are shown in Fig. \ref{fig:c_tauPredict}(c).

In Fig. \ref{fig:c_tauPredict}(a), we see a max $c_{\tau}$ of 2.53 Nmm/$^\circ$ when rotation is 9.7$^\circ$ for the blocked force path. This tends towards a value of 0.31 Nmm/$^\circ$ when the HSA rotation is 117.5$^\circ$. The cylinder prediction path starts at -0.58 Nmm/$^\circ$ at 9.7$^\circ$, reaches a maximum of 0.72 Nmm/$^\circ$ at 48.6$^\circ$ and reaches 0.31 Nmm/$^\circ$ at 117.5$^\circ$. The values for blocked force are always positive as the torque values start at zero and then steadily climb. For the cylinder prediction path, there is a non-zreo starting displacement. This causes the values to start negative as the HSA applies a negative torque, before transitioning to a positive value.

\begin{figure}
    \centering
    \includegraphics[width=0.48\textwidth]{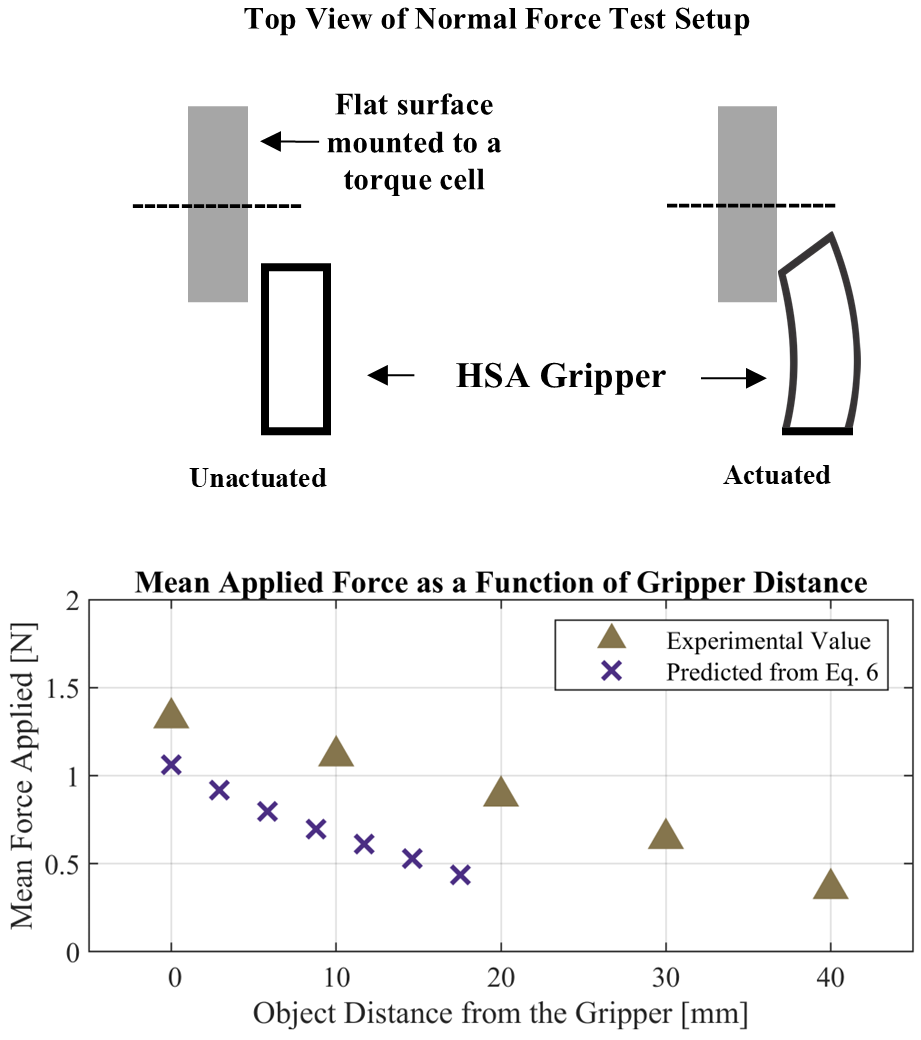}
    \caption{This figure represents the measured normal force as a function of object distance for a single HSA finger at 120$^\circ$ HSA rotation, compared to the predicted normal force from measured $c_{\tau}$ values. The top figure shows the 2D representation of the test setup used to measure the normal force applied to an object at different object-to-gripper distances. Here, we measure the torque applied on the flat surface for an interval of 10 seconds and calculate the applied normal force.}
    \label{fig:f_nPredict}
\end{figure}

In Fig. \ref{fig:c_tauPredict}(b), we see a larger maximum $c_{\tau}$ of 2.77 Nmm/$^\circ$ when rotation is 9.8$^\circ$ for the blocked force path of the long HSA. This tends towards a value of 0.21 Nmm/$^\circ$ when the HSA rotation is 118.6$^\circ$. This is considerably smaller than the value computed for the short HSA. The cylinder prediction path starts at a larger value of -1.58 Nmm/$^\circ$ at 9.8$^\circ$, reaches a maximum of 0.58 Nmm/$^\circ$ at a later 57.6$^\circ$ and reaches 0.21 Nmm/$^\circ$ at 118.6$^\circ$. Since the Cylinder force characterization test uses a motor input of 120$^\circ$, a $c_{\tau}$ value of 0.21 is selected. Thus a full value for the prediction can be computed. This is shown in Fig. \ref{fig:instronResults} as a black dashed line. In Fig. \ref{fig:f_nPredict}, we see the mean applied normal force as a the gripper to object distance increases. To measure this data, we used a flat surface to apply the normal force which was fixed to a torque cell. We then recorded the torque at different object to gripper distances for a period of 10 seconds. The mean applied force was then calculated using the recorded torque data. Additionally, we predicted the normal force using Eq. \ref{eq:forceG+HSA} with the corresponding $c_{\tau}$ values at different finger extensions as shown in Fig. \ref{fig:f_nPredict}. Here, we consider the extension values that were used for characterizing the short HSA, and correlate the measured $c_{\tau}$ values to predict normal force $F_n$ applied by a single finger. The predicted values were calculated at 0, 2.9, 5.8, 8.7, 11.6, 14.6 and 17.5 mm of HSA extension, and the measured values were at 0, 10, 20, 30 and 40 mm of finger extension. Based on our prediction, the values show a similar trend to the measured values, with a faster decay for the predicted values.

we can see that the HSA gripper is capable of achieving a peak holding force of 5.8 N using pinch grasp, and 14.6 N with planar caging grasp, from the cylinder holding force test shown in Fig. \ref{fig:instronResults}. Over the first 3 mm, we observe a grasp stiffness of 1.64 N/mm with pinch grasp, and a grasp stiffness of 3.06 N/mm with the planar caging grasp. The prediction model under represents the contribution from the HSA, predicting a value of 1.47 N when the realized performance is 1.98 N. The slope of the dashed line represents the contribution from the TR-SLL element. Based on Fig. \ref{fig:failureModes}, the minimum set of slipping, twisting, and shearing will govern the maximum holding force the gripper can sustain. Since the slope of the pinch grasp is shallower than the predicted value, this demonstrates that the gripper is not limited by the twisting failure condition. By visual inspection from the supplementary video, the slip condition is insufficient normal force. It can also be observed that the initial holding force using pinch grasp reaches a maximum of 2.31 N, whereas the initial holding force using the planar caging grasp gets to a maximum of 0.47 N. This is due to the highest amount of torsional stiffness $c_{\tau}$ at the distal end of the HSA finger, which significantly reduces as the object contact point moves towards the proximal end of the finger. With respect to the peak holding force, planar caging grasp provides higher performance in lifting larger payloads with a peak holding force of 14.6 N since this grasping mode utilizes the palm to encompass the object with three surface contact points as shown in Fig. \ref{fig:instronResults}.

\subsection{YCB Object Grasping}

We demonstrate the TR-SLL gripper's ability to pickup objects from the YCB object dataset \cite{YCBObjects}. We chose to select objects that fit within the gripper (0 to 105 mm). Due to the torque from the HSA, the grippers curve in a smile shape. This causes a minimum height of object below which the object cannot be grasped perpendicular to the gravity field. This height was measured at 14.5 mm. Thus, the set was limited to objects that were at least 14.5 mm tall when standing in their default orientation and able to fit between the opened jaws of the gripper. This left us with 43 objects. All objects were filled to match their weight and approximate mass distribution according to the data from \cite{YCBObjects}.

Within the object dataset, we chose three possible pick states: Success, Partial Success, and Failure. A successful lift is fully lifting the object off the ground and maintaining it in the grasp for at least 30 seconds. A partial success is lifting the object but not maintaining contact for the full time duration. A failure is not successfully maintaining contact on the object. Of the objects that were not successful in the pinch grasp, a caging grasp was attempted. The maximum of these two grasp attempts is reported as the number of successes, partial successes, and failures. All tests were done using a bang-bang controller switching between two states: open (0$^\circ$) and closed (120$^\circ$). The motors used were QDD100 Beta 3 servos from MJbots. The default gains were used on the motors. Across both pinch and caging grasps, the success rate of the gripper was 86\% (37/43), the partial success rate was 9.3\% (4/43), and 4.7\% (2/43) were unsuccessful.

Object from the YCB data set were first tested using a pinch grasp. The HSA gripper was able to successfully lift 32 objects, with a success rate of 74.41\%. The 32 successful objects were the chips can, the cracker box, the mustard bottle, the gelatin box, pudding box, potted meat can, banana, strawberry, apple, lemon, peach, pear, orange, plum, the wine glass, the mug, spatula, padlock, flat and Phillips screwdrivers, the soft ball, baseball, tennis ball, racquetball, golf ball, the smallest and the largest cup, foam brick, colored wood block, the nine-hold peg test, the timer, and the Rubik’s Cube. Out of the ten cups, we tested the extreme sizes (smallest and largest) to ensure objects had similar importance within the dataset.

There were five partial successes: the sugar box, tomato soup can, master chef can, the wood block, and the large marker. In this case, while the gripper did not experience any twisting load during object pickup, the first 4 objects listed were partially successful due to insufficient normal force. This corresponds to the slip failure mode in soft grippers. The large marker was a partial success as contact on the object was on the top half of the cylinder, leading to an eventual drop. Hence, 11.6\% of the total objects were partially successful with the pinch grasp.

Six objects from the set were failures: the hammer, the windex bottle, bleach cleanser, the nine-hold peg test, the die, and the power drill. The mode of failure was either insufficient normal force or shearing of the Ecoflex layer.

A planar caging grasp was used on the objects that were not in the previous success category. Objects were caged using the palm. Out of eleven objects that were tested with the planar caging grasp, five were successful (nine-hole peg test, master chef can, wood block, sugar box, power drill), two were partially successful (windex bottle, bleach cleanser), and four were failures (hammer, die, large marker, tomato soup can). Two of the objects (windex bottle, bleach cleanser) were partially successful due to insufficient normal force. Two of the four objects (hammer, large marker) once again failed due to insufficient normal force while two (soup can, die) automatically failed due to the objects being unable to reach both the gripper and the palm.

\begin{figure}
    \centering
    \includegraphics[width=0.48\textwidth]
    {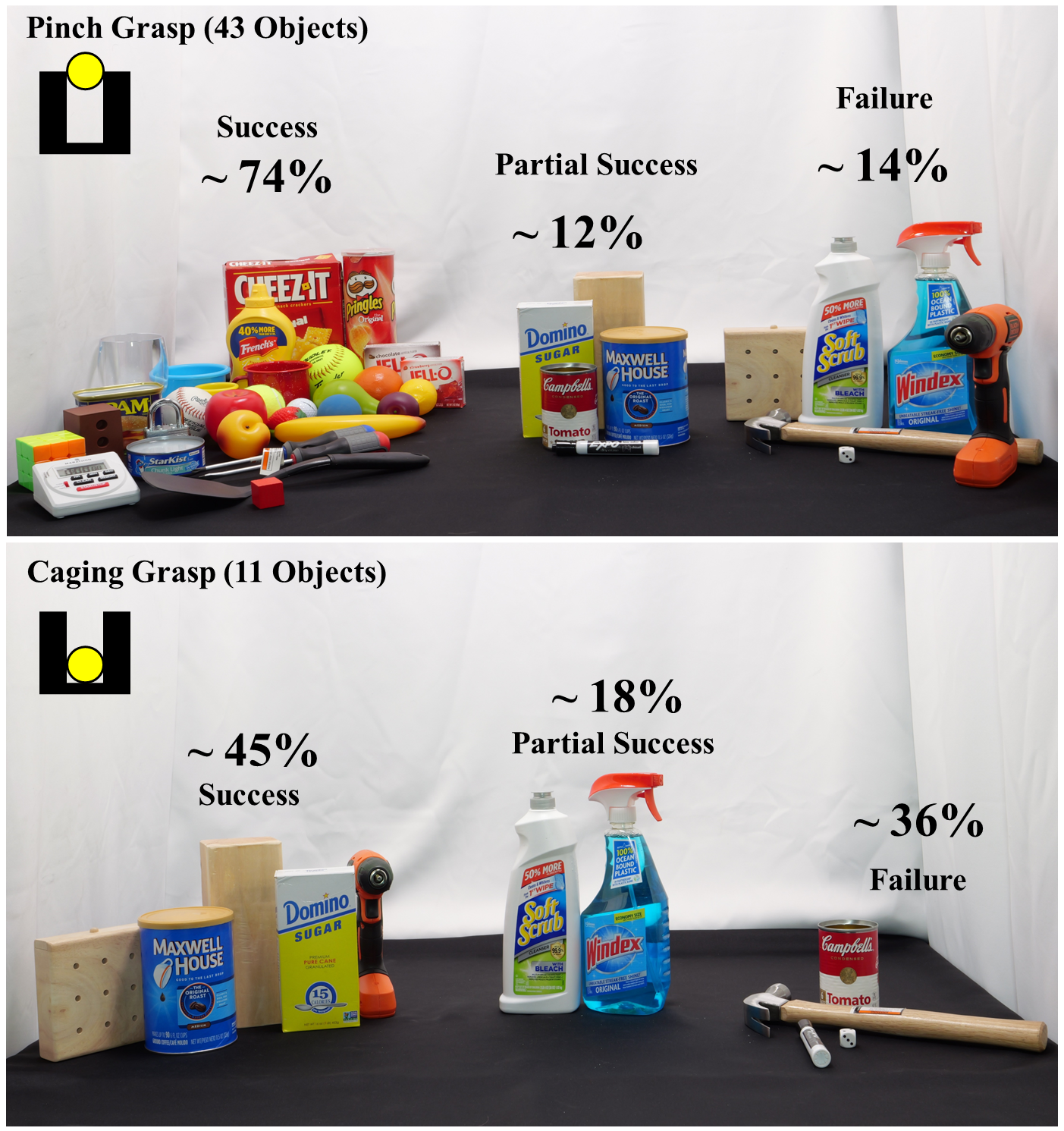}
    \caption{This figure shows the YCB test objects we evaluated the HSA gripper against. Objects were filtered to have a critical dimension that fit within the jaws of the gripper. First a pinch grasp was performed on the dataset. A success was defined as lifting the object and holding it for 30 seconds. A partial success was fulfilled if the object was able to be picked up but was not stable after 30 seconds. If the object was unable to be lifted off the ground, it was a failure. Of the objects that were not fully stable after the time period, a caging grasp was employed. The same success metrics were employed. The gripper was successfully able to lift all objects across both grasp modes. However, three objects only achieved a peak of partial success.}
    \label{fig:YCB Objects}
\end{figure}

The overall unsuccessful objects for both the pinch and planar caging grasps accounted to only 13.9\% of the entire tested objects. The primary failure modes were determined to be slipping and material shearing. This shows that the TR-SLL increased the gripper's resistance to torsion.
\section{Conclusion and Future Work}


In this paper, we developed a torsion resistant strain limiting layer (TR-SLL) to increase payload capacity by increasing torsional resistance. A comprehensive design study was conducted using FEA simulations to understand the design landscape of the torsion resistant strain limiting layer (TR-SLL). The number of triangles on a 102-mm long strain limiting layer was varied to characterize the performance of the TR-SLL in plane and out of plane. It was found that designs near equilateral triangles best resist torsional loads. We also characterized the 3D force and torque landscape for HSAs in terms of displacement and rotation. This data was then used to predict the payload capacity due to twisting of the gripper and found that twist was not the dominant failure mode for the gripper. Combining the TR-SLL with HSAs enables gripper construction from a single HSA, enabling more nimble, and performant grasping.

This work found that adding a torsion resistant strain limiting layer dramatically increased torsional performance, demonstrating a grasp stiffness value of 3060 N/m. The addition of the TR-SLL allowed a electrically-driven handed shearing auxetic gripper to pick up a 5 kg dumbbell, maxing out the payload capacity of the robot arm it was installed on. The TR-SLL gripper was also tested on a subset of the YCB data set and successfully picked up 86\% of the objects. Future work for these TR-SLL grippers include sensorization, additional characterization of the TR-SLL design space to incorporate additional design parameters, and more intelligent controls to further improve grasp performance. This work lays the foundation to easily incorporate TR-SLLs into existing soft robots and demonstrates additional payload capacity that is often lost due to torsional deflection in soft grippers.
\section*{Acknowledgment}

The authors would like to thank Josh Crumrine and Joe Torky for their help with 3D printing, and David Oh for providing python scripts to program the servos.

\bibliographystyle{IEEEtran}
\bibliography{hsa}
\end{document}